\documentclass[10pt,twocolumn,letterpaper]{article}

\usepackage{cvpr}
\usepackage[utf8]{inputenc} % allow utf-8 input
\usepackage[T1]{fontenc}    % use 8-bit T1 fonts
\usepackage{url}            % simple URL typesetting
\usepackage{booktabs}       % professional-quality tables
\usepackage{amsfonts}       % blackboard math symbols
\usepackage{nicefrac}       % compact symbols for 1/2, etc.
\usepackage{microtype}      % microtypography
\usepackage{times}
\usepackage{graphicx,xspace} % more modern
\usepackage[dvipsnames]{xcolor}
\usepackage{epstopdf}
\usepackage{subfigure}
\usepackage{paralist}
\usepackage{amsmath,amssymb,amsthm,array,amsfonts,mathtools}

\usepackage{algorithm}
\usepackage{algorithmic}

\usepackage{booktabs,multirow,rotating}
\graphicspath{{./}{./figures/}}
\DeclareGraphicsExtensions{.pdf,.jpeg,.png,.jpg}

\makeatletter
\DeclareRobustCommand\onedot{\futurelet\@let@token\@onedot}
\def\@onedot{\ifx\@let@token.\else.\null\fi\xspace}

\def\eg{\emph{e.g}\onedot} 
\def\ie{\emph{i.e}\onedot}

\def\etal{\emph{et al}\onedot}
\makeatother

\def\Vec#1{{\boldsymbol{#1}}}
\def\Mat#1{{\boldsymbol{#1}}}

\usepackage[pagebackref=true,breaklinks=true,letterpaper=true,colorlinks,bookmarks=false]{hyperref}

\cvprfinalcopy % *** Uncomment this line for the final submission

\def\cvprPaperID{2113} % *** Enter the CVPR Paper ID here

\ifcvprfinal\fi
\begin{document}

%%%%%%%%% TITLE
\title{End-to-End Learning of Motion Representation for Video Understanding}
\author{Lijie Fan\thanks{indicates equal contributions. This work was conducted when Lijie Fan was served as a research intern in Tencent AI Lab.}\ $^{2}$, Wenbing Huang$^{\ast1}$, Chuang Gan$^{3}$, Stefano Ermon$^{4}$, Boqing Gong$^{1}$, Junzhou Huang$^{1}$\\
$^1$ Tencent AI Lab,
$^2$ Tsinghua University, Beijing, China\\
$^3$ MIT-Watson Lab,
$^4$ Department of Computer Science, Stanford University \\
{\tt\small flj14@mails.tsinghua.edu.cn, hwenbing@126.com, ganchuang1990@gmail.com}\\
{\tt\small ermon@cs.stanford.edu, boqinggo@outlook.com, jzhuang@uta.edu}
%\and
}
\maketitle
\thispagestyle{empty}

%%%%%%%%% ABSTRACT
\begin{abstract}
Despite the recent success of end-to-end learned representations, hand-crafted optical flow features are still widely used in video analysis tasks.
To fill this gap, we propose TVNet, a novel end-to-end trainable neural network, to learn optical-flow-like features from data.
TVNet subsumes a specific optical flow solver, the TV-L1 method, and is initialized by unfolding its optimization iterations as neural layers.
TVNet can therefore be used directly without any extra learning. Moreover, it can be naturally concatenated with other task-specific networks to formulate an end-to-end architecture, thus making our method more efficient than current multi-stage approaches by avoiding the need to pre-compute and store features on disk.
Finally, the parameters of the TVNet can be further fine-tuned by end-to-end training.
This enables TVNet to learn richer and task-specific patterns beyond exact optical flow.
Extensive experiments on two action recognition benchmarks verify the effectiveness of the proposed approach.  Our TVNet achieves better accuracies than all compared methods, while being competitive with the fastest counterpart in terms of features extraction time.
\end{abstract}

%%%%%%%%% BODY TEXT

\section{Introduction}
\label{Sec:introduction}

%Analyzing human actions from videos is a highly active line of research due to its broad applications such as video surveillance~\cite{poppe2010survey} and human behavior analysis~\cite{Ji_TPAMI13}.
%Although deep architectures and in particular Convolutional Neural Networks (CNNs) have significantly improved image-based tasks (\eg image classification~\cite{he2016deep} and objective detection~\cite{ren2015faster}), the progress on video analysis is still far from satisfactory, which merely reflecting the difficulty associated with analyzing spatiotemporal data. In fact, applying deep solutions demands a particular network design for modeling the motion patterns.
Deep learning and especially Convolutional Neural Networks (CNNs) have revolutionized image-based tasks, \eg, image classification~\cite{he2016deep} and objective detection~\cite{ren2015faster}. However, the progress on video analysis is still far from satisfactory, reflecting the difficulty associated with learning representations for spatiotemporal data. We believe that the major obstacle is that the distinctive motion cues  in videos demand some new network designs, which are yet  to be found and tested.

\begin{figure}[!ht]
\vskip -0.2 in
\begin{center}
\subfigure{
\includegraphics[width=7.5cm]{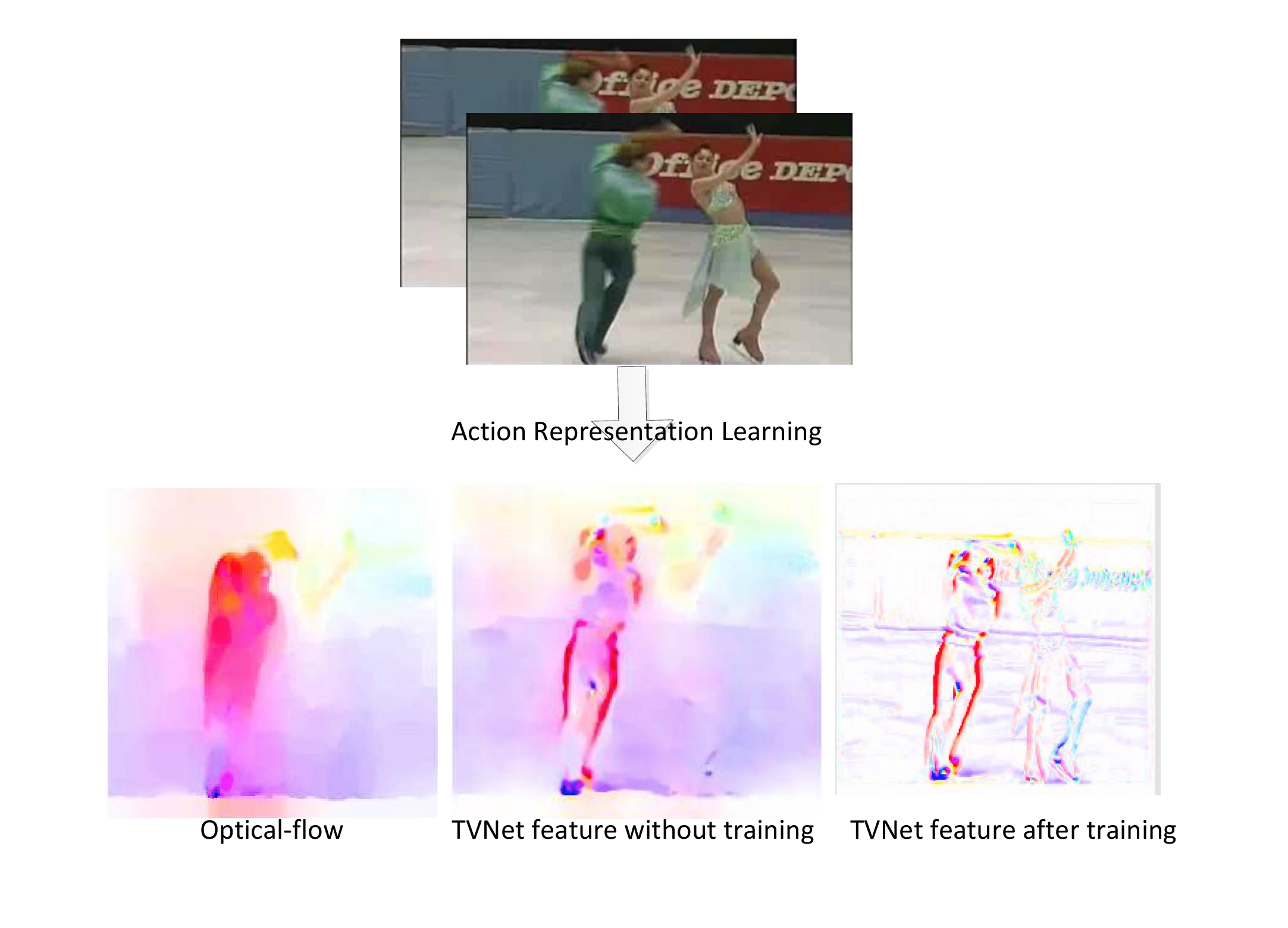}
}
\vskip -0.2 in
\caption{Visualization results of optical-flow-like motion features by TV1~\cite{zach2007duality}, TVNet (without training) and TVNet (with training).}
\label{Fig:intro}
\end{center}
\vskip -0.2 in
\end{figure}

While there have been some attempts ~\cite{tran2015learning} to learn features by convolution operations over both spatial and temporal dimensions, optical flow is still widely and effectively used for video analysis~\cite{Ng_CVPR15,Feichtenhofer_NIPS16,Feichtenhofer_CVPR16,Wang_ECCV16,gkioxari2015finding,peng2016multi}.
The optical flow, as the name implies, captures the displacements of pixels between two consecutive frames~\cite{zach2007duality}. Thus,
applying the optical flow to the video understanding tasks enables one to model the motion cues explicitly, conveniently, but inefficiently. It is often computationally expensive to estimate the optical flows.
%The work by~\cite{tran2015learning} proposed to model spatiotemporal patterns by 3-Dimensional Convolutional Networks (3D ConvNets). Such networks model appearance and motion by performing the convolution operations over the spatial and temporal directions simultaneously. Another line of researches (\eg the two-stream networks~\cite{Simonyan_NIPS14}) resorts to analyze motion clues by first extracting optical flow data from the raw RGM images and then training a CNN on the extracted flow data.
%The optical flow, as the name implies, is defined as the displacements of pixels between to continuous frames~\cite{zach2007duality}. Thus,
%applying the optical flow data as the action representor is capable of modeling the motion clues explicitly and thus efficiently.
A currently successful example of applying optical flow to video understanding is the two-stream model~\cite{Simonyan_NIPS14}, where a CNN is trained on the optical flow data to learn action patterns.
Various extensions of the two-stream model have been proposed and achieved state-of-the-art results on serval tasks including action recognition \cite{Ng_CVPR15,Feichtenhofer_NIPS16,Feichtenhofer_CVPR16,Wang_ECCV16} and action detection~\cite{gkioxari2015finding,peng2016multi}.

Despite the remarkable performance, current optical-flow-based approaches have notable drawbacks:
\begin{itemize}
  \item \textbf{Training is a two-stage pipeline.} In the first stage, the optical flow for every two consecutive frames is extracted via the optimization-based method (\eg TV-L1~\cite{zach2007duality}). In the second stage, a CNN is trained on the extracted flow data. These two stages are separated and the information (\eg gradients) from the second stage cannot be used to adjust the process of the first stage.
  \item \textbf{Optical flow extraction is expensive in space and time.} The extracted optical flow has to be written to the disk for both the training and testing. For the UCF-101 dataset~\cite{soomro2012ucf101} which contains about 10 thousands videos, extracting optical flows for all data via the TV-L1 method  takes one GPU-day, and storing them costs more than one TeraByte of storage for the original fields as floats (often a linear JPEG normalization is required to save storage cost~\cite{Simonyan_NIPS14}).
\end{itemize}

To tackle the above mentioned problems, we propose a novel neural network design for learning optical-flow like features in an end-to-end manner. This network, named TVNet, is obtained by imitating and unfolding the iterative optimization process of TV-L1~\cite{zach2007duality}. In particular, we formulate the iterations in the TV-L1 method as customized layers of a neural network. As a result, our TVNet is well-grounded and can be directly used without additional training by any groundtruth optical flows.
%First, the resulting formula combines the expressive power of a conventional deep network with the internal structure of the model-based approach, while allowing inference to be performed in a fixed number of layers that can be optimized for best performance.

Furthermore, our TVNet is end-to-end trainable, and can therefore
%is directly conducted on RGB images
be naturally connected with a tasks-specific network (\eg action classification network) to form a ``deeper'' end-to-end trainable architecture. As a result, it is not necessary to pre-compute or store the optical-flow features anymore.
%is not necessary anymore.

Finally, by performing the end-to-end learning, it is possible to
%our network is able to be transformed to a trainable version by relaxing certain variables. The variable relaxation strick enables the network to further
fine-tune the weights of the TVNet that is initialized as a standard optical flow feature extractor. This allows us to discover richer and task-specific features (compared to the original optical flow) and thus to deliver better performance.

To verify the effectiveness of the proposed architecture, we perform experimental comparisons between the proposed TVNet and several competing methods on two action recognition benchmarks (HMDB51~\cite{kuehne2011hmdb} and UCF101~\cite{soomro2012ucf101}).% and one action similarity detection task (ASLAN~\cite{kliper2012action}).

%To sum up, the contributions of this paper are as follow:
%\begin{itemize}
%  \item We develop a novel neural network to learn action representation by unfolding the iterations of the TV-L1 method as particular neural layers. The dubbed network, \ie TVNet is well-initialize, end-to-end and trainable.
%  \item Although initialized as a specific TV-L1 architecture, the proposed TVNet can be further finetuned to learn richer and more task-oriented action features other than the optical flow fields.
%  \item Our TVNet achieves better accuracies than other action representation counterparts (\eg, TV-L1~\cite{zach2007duality}, FlowNet2.0~\cite{ilg2016flownet}) and 3D Convnets~\cite{tran2015learning} on two action recognition benchmarks (HMDB51~\cite{kuehne2011hmdb} and UCF101~\cite{soomro2012ucf101}) and one action similarity detection task (ASLAN~\cite{kliper2012action}).
%\end{itemize}

To sum up, this paper makes the following contributions:
\begin{compactitem}
  \item We develop a novel neural network to learn motions from videos by unfolding the iterations of the TV-L1 method to customized neural layers. The network, dubbed TVNet, is well-initialized and end-to-end trainable.
  \item Despite being initialized as a specific TV-L1 architecture, the proposed TVNet can be further fine-tuned to learn richer and more task-oriented features than the standard optical flow.
  \item Our TVNet achieves better accuracies than other action representation condunterparts (\eg, TV-L1~\cite{zach2007duality}, FlowNet2.0~\cite{ilg2016flownet}) and 3D Convnets~\cite{tran2015learning} on the two action recognition benchmarks, \ie,72.6\% on HMDB51 and 95.4\% on UCF101. %, and 79.2\% on ASLAN.
\end{compactitem}

\section{Related Work}
\label{Sec:related-work}
Video understanding, such as action recognition and action similarity detection, has attracted a lot of research attention in the past decades.
Different from static image understanding, video understanding requires more reliable motion features to reflect the dynamic changes occurring in videos. Laptev \etal~\cite{laptev2005space} proposed a spatio-temporal interest points (STIPs) method by extending Harris corner detectors to 3-dimensional space to capture motion. Similarly, the 3D extensions of SIFT and HOG have also been investigated~\cite{Mosift} and ~\cite{klaser2008spatio}, respectively. Wang \etal~\cite{wang2013action} proposed improved Dense Trajectories (iDT), where the descriptors were obtained by tracking densely sampled points and describing the volume around the tracklets by histograms of optical flow (HOF) and motion boundary histograms (MBH). Despite its stat-of-the-art performances, IDT is computationally expensive and becomes intractable on large-scale video dataset.

Motivated by the promising results of deep networks on image understanding tasks, there have also been a number of attempts to develop
deep architectures to learn motion features for video understanding~\cite{Ji_TPAMI13,Karpathy_CVPR14,Ng_CVPR15,DevNet, Herath_IVC2017,zhang2016real,gan2016you,gan2016webly}.
The leading approaches fall into two broad categories. The first one is to learn appearance and motion jointly by extending 2D convolutional layers to 3D counterparts ~\cite{tran2015learning,Ji_TPAMI13}, including recently proposed I3D~\cite{carreira2017quo} and P3D~\cite{qiu2017learning}. However, modeling motion information through 3D convolutional filters is computationally expensive, and large-scale training videos are needed for desired performance~\cite{carreira2017quo}.
The other category of work is based on two-stream networks~\cite{Simonyan_NIPS14,Ng_CVPR15,Wang_ECCV16,Feichtenhofer_NIPS16,Feichtenhofer_CVPR16,long2017attention}. %They propose to learn spatiotemporal features in a separated fashion.
This line of approaches trains two networks, one using the appearance (\ie, RGB) data and the other one using hand-crafted motion features such as optical flow to represent motion patterns.
In contrast, in our method, the motion descriptor is learned with a trainable neural network rather than hand-crafted. As a consequence, our optical-flow-like motion features can be jointly learned and fine-tuned using a task-specific network. Additionally, we do not need to store and read the optical flow from disk, leading to significant computational gains.
%which significantly saves the computational cost.

A recent research topic is to estimate optical flow by CNNs~\cite{dosovitskiy2015flownet,teney2016learning,ranjan2016optical,ilg2016flownet,long2017attention,bian2017revisiting}. These approaches cast the optical flow estimation as an optimization problem with respect to the CNN parameters. A natural idea is to combine the flow CNN with the task-specific network to formulate an end-to-end model (see for example in~\cite{zhu2017hidden}). Nevertheless, an obvious issue of applying the flow nets is that they require thousands of hundreds of groundtrue flow images to train the parameters of the flow network to produce meaningful optical flows (see~\cite{dosovitskiy2015flownet}). For real applications, it is costly to obtain the labeled flow data. In contrast, our network is well initialized as a particular TV-L1 method and is able to achieve desired performance even in its initial form (without fine-tuning).

Recently, Ng \etal \cite{ng2016actionflownet} proposed to train a single stream convolutional neural network to jointly estimate
optical flow and recognize actions, which is most relevant to our work. To capture the motion feature, they formulated FlowNet~\cite{fischer2015flownet} to learn the optical flow from %the
%synthesis
synthetic
ground truth data. Though the results are promising, the approach still lags behind the state of the arts in terms of accuracy compared to traditional approaches. This is due to the well known gap between synthetic and real videos. Contrastly, our network is formulated by unfolding the TV-L1 method that has been applied successfully to action recognition and we do not rely on the groundtruth of optical flow for training. Thus, our network combines the strengths of both TV-L1 and deep learning.

\section{Notations and background}
%We review the TV-L1 method in this section.
% (a)
%\s{definite the video as a sequence of images?}

\subsection {Notations}
A video sequence can be written as a function of three arguments, $\Mat{I}_t(x,y)$, where $x,y$ index the spatial dimensions $t$ is for the time dimension. Denote by $\Omega$ all the coordinates of the pixels in a frame. The function value $\Mat{I}_t(x,y)$ corresponds to the pixel brightness at position $\Vec{x}=(x,y)\in\Omega$ in the $t$-th video frame. A point $\Vec{x}$ may move from time to time across the video frames, and the optical flow is to track such displacement between adjacent frames. We denote by $\Vec{u}^t(\Vec{x})=(u^t_1(\Vec{x}), u^t_2(\Vec{x}))$ the displacement of the point $\Vec{x}$ from time $t$ to the next frame $t+1$. We omit the superscript $t$ and/or argument $\Vec{x}$ from $\Vec{u}^t(\Vec{x})$ when no ambiguity is caused.

\begin{figure*}[!th]
\vskip -0.2 in
\begin{center}
\subfigure{
\includegraphics[width=18cm,height=6cm]{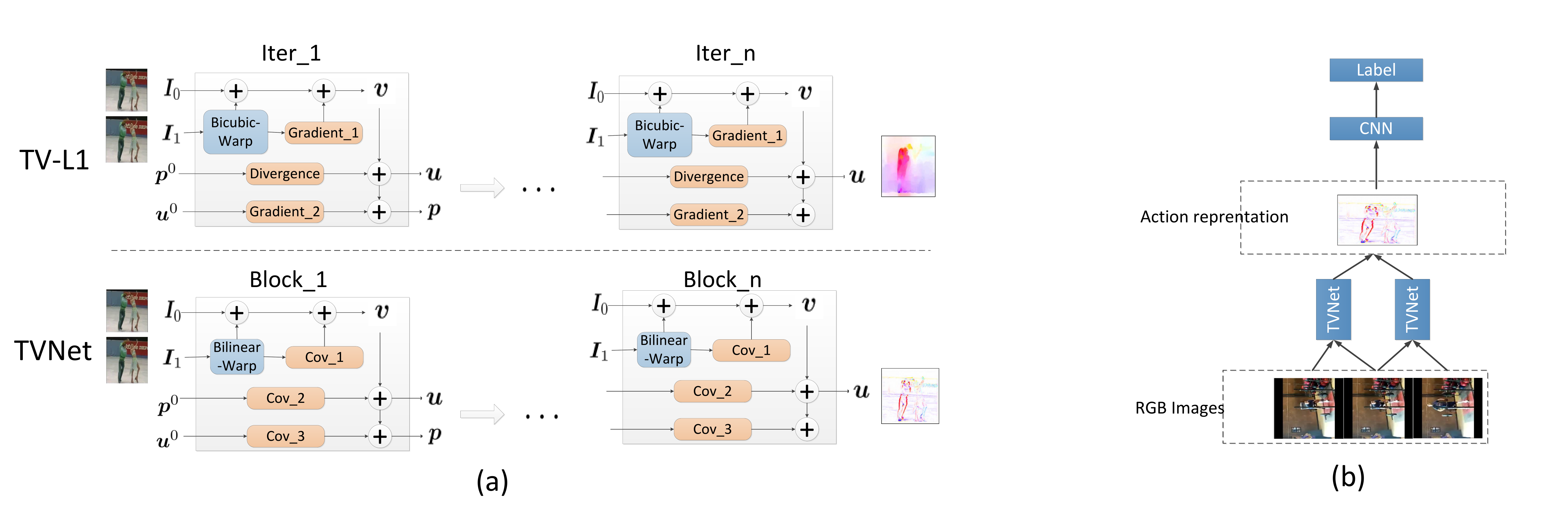}
}
\vskip -0.1 in
\caption{(a) Illustration of the process for unfolding TV-L1 to TVNet. For TV-L1, we illustrate each iteration of Algorithm~\ref{Alg:tvl1}. We reformulate the bicubic warping, gradient and divergence computations in TV-L1 to bilinear warping and convolution operations in TVNet. (b) The end-to-end model for action recognition. }
\label{Fig:tvnet}
\end{center}
\vskip -0.2 in
\end{figure*}

\subsection{The TV-L1 method}
\label{Sec:tvl1}
Among the existing approaches to estimating  optical flows, the TV-L1 method~\cite{zach2007duality} is especially appealing for its good balance between efficiency and accuracy. We review it in detail in this subsection to make the paper self-contained. The design of our TV-Net (cf.\ Section~\ref{Sec:tvnet}) is directly motivated by the optimization procedure of TV-L1.

The main formulation of TV-L1 is as follows,
\vskip -0.25 in
\begin{eqnarray}
\label{Eq:energy-function}
\min_{\Mat{u}(\Vec{x}),\Vec{x}\in\Omega} \sum_{\Vec{x}\in \Omega} (|\nabla \Vec{u}_1(\Vec{x})|+|\nabla \Vec{u}_2(\Vec{x})|) + \lambda|\rho(\Vec{u}(\Vec{x}))|,
\end{eqnarray}
\vskip -0.1 in \noindent
where the first term $|\nabla \Vec{u}_1|+|\nabla \Vec{u}_2|$ accounts for the \emph{smoothness condition}, while the second term $\rho(\Vec{u})$ corresponds to the famous \emph{brightness constancy assumption}~\cite{zach2007duality}. In particular, the brightness of a point $\Vec{x}$ is assumed to remain the same after it shifts to a slightly different location in the next frame, i.e., $\Mat{I}_0(\Vec{x}+\Vec{u})\approx \Mat{I}_{1}(\Vec{x})$. Accordingly, $\rho(\Vec{u})=\Mat{I}_{1}(\Vec{x}+\Vec{u})-\Mat{I}_0(\Vec{x})$ is defined in order to penalize the brightness difference in the second term.
Since the function $\Mat{I}_{1}(\Vec{x}+\Vec{u})$ is highly non-linear with respect to $\Vec{u}$, Zach et al.~\cite{zach2007duality} approximate the brightness difference $\rho(\Vec{u})$ by the Taylor expansion at an initial displacement $\Vec{u}^{0}$,  leading to $\rho(\Vec{u})\approx\nabla\Mat{I}_{1}(\Vec{x}+\Vec{u}^{0})(\Vec{u}-\Vec{u}^{0}) + \Mat{I}_{1}(\Vec{x}+\Vec{u}^0) -\Mat{I}_0(\Vec{x})$.

The above gives a first-order approximation to the original problem and linearizes it to an easier form. Furthermore, the authors introduce an auxiliary variable $\Vec{v}$ to enable a convex relaxation of the original problem
\vskip -0.25 in
\begin{eqnarray}
\label{Eq:convex-relaxation}
\min_{\{\Mat{u},\Mat{v}\}} \sum_{\Vec{x}\in \Omega} (|\nabla \Vec{u}_1|+|\nabla \Vec{u}_2|) + \frac{1}{2\theta}|\Vec{u}-\Vec{v}|^2 + \lambda|\rho(\Vec{v})|,
\end{eqnarray}
\vskip -0.1 in \noindent
in which a very small $\theta$ can force $\Vec{u}$ and $\Vec{v}$ to be equal at the minimum. This objective is minimized by alternatively updating $\Vec{u}$ and $\Vec{v}$. The details of the optimization process are presented in Algorithm~\ref{Alg:tvl1}, where the variables $\Vec{p}_1$ and $\Vec{p}_2$ are the dual flow vector fields.

\begin{algorithm}[t]
\caption{The TV-L1 method for optical flow extraction.}
\label{Alg:tvl1}
\begin{algorithmic}
   \STATE {\bfseries Hyper-parameters:} $\lambda,\theta,\tau,\epsilon,N_{warps},N_{iters}$
   \STATE {\bfseries Input:} $\Mat{I}_0,\Mat{I}_1,\Vec{u}^0$
   \STATE $\Vec{p}_1=[\Vec{p}_{11},\Vec{p}_{12}]=[\Vec{0},\Vec{0}]$;
   \STATE $\Vec{p}_2=[\Vec{p}_{21},\Vec{p}_{22}]=[\Vec{0},\Vec{0}]$;
   \FOR{$w=1$ {\bfseries to} $N_{warps}$}
   \STATE Warp $\Mat{I}_1(\Vec{x}+\Vec{u}^0), \nabla\Mat{I}_1(\Vec{x}+\Vec{u}^0)$ by interpolation;
   \STATE $\rho(\Vec{u})=\nabla\Mat{I}_1(\Vec{x}+\Vec{u}^{0})(\Vec{u}-\Vec{u}^{0})+\Mat{I}_1(\Vec{x}+\Vec{u}^0)-\Mat{I}_0(\Vec{x})$, $n=0$;
   \WHILE{$n<N_{iters}$ and ${stopping\_criterion}>\epsilon$}
   \STATE $\Vec{v}= \left\{
   \begin{array}{ll}
   \lambda\theta\nabla\Mat{I}_1 & \rho(\Vec{u})<-\lambda\theta|\nabla\Mat{I}_1|^2,\\
   -\lambda\theta\nabla\Mat{I}_1 & \rho(\Vec{u})>\lambda\theta|\nabla\Mat{I}_1|^2,\\
   -\rho(\Vec{u})\frac{\nabla\Mat{I}_1}{|\nabla\Mat{I}_1|^2} & \text{otherwise},
   \end{array}
   \right. $ where $\nabla\Mat{I}_1$ represents $\nabla\Mat{I}_1(\Vec{x}+\Vec{u}^0)$ for short;
   \STATE $\Vec{u}_d=\Vec{v}+\theta \mathrm{div}(\Vec{p}_d)$, $d=1,2$;
   \STATE $\Vec{p}_d=\frac{\Vec{p}_d+\tau/\theta \nabla\Vec{u}_d}{1+\tau/\theta|\nabla\Vec{u}_d|}$, $d=1,2$;
   \STATE $n=n+1$;
   \ENDWHILE
   \ENDFOR
\vskip -0.1in
\end{algorithmic}
\end{algorithm}

\paragraph{Understanding Algorithm~\ref{Alg:tvl1}.} The core computation challenge in the algorithm is on the pixel-wise computations of  the \emph{gradients} (\ie, $\nabla\Mat{I}_1$ and $\nabla\Vec{u}_d$), \emph{divergence} (\ie, $div(\Vec{p})$), and \emph{warping} (\ie, $\Mat{I}_1$ and $\nabla\Mat{I}_1$).
%In particular, The central and forward differences are adopted to compute the gradients $\nabla\Mat{I}_1(\Vec{x}+\Vec{u}^0)$ and $\nabla\Vec{u}_d$, respectively; the backward difference is applied to compute the divergence $\mathrm{div}(\Vec{p}_d)$; the bicubic interpolation [] is applied to warp the image $\Mat{I}_1$ by a flow field $\Vec{u}^0$.
The details of the numerical estimations are provided as below.
%Note that the Neumann boundary condition is used for the boundary points.
\begin{itemize}
  \item \textbf{Gradient-1.} The gradient of the image $\Mat{I}_1$ is computed by central difference:
  \vskip -0.25 in
  \begin{eqnarray}
  \label{Eq:central}
  \frac{\partial \Mat{I}_1(i,j)}{\partial x} = \left\{
   \begin{array}{ll}
   \frac{\Mat{I}_1(i+1,j)-\Mat{I}_1(i-1,j)}{2} & 1<i<W,\\
   0 & \text{otherwise}.
   \end{array}
   \right.
  \end{eqnarray}
  \vskip -0.1 in \noindent
  We can similarly compute $\frac{\partial}{\partial y} \Mat{I}_1(i,j)$ along the $j$ index.
  \item \textbf{Gradient-2.} The gradient of each component of the flow $\Mat{u}$ is computed via the forward difference:
  \vskip -0.25 in
  \begin{eqnarray}
  \label{Eq:forward}
  \frac{\partial \Vec{u}_d(i,j)}{\partial x} = \left\{
   \begin{array}{ll}
   \Vec{u}_d(i+1,j)-\Vec{u}_d(i,j) & 1\leq i<W,\\
   0 & \text{otherwise},
   \end{array}
   \right.
  \end{eqnarray}
  \vskip -0.1 in \noindent
  where $d\in\{1,2\}$. Also, $\frac{\partial}{\partial y} \Mat{u}_d(i,j)$ can be similarly computed by taking the difference on the $j$ index.
  \item \textbf{Divergence.} The divergence of the dual variables $\Vec{p}$ is computed via the backward difference:
  \vskip -0.25 in
  \begin{eqnarray}
  \label{Eq:backward}
   \nonumber
   && \mathrm{div}(\Vec{p}_d)(i,j) \\
   \nonumber
   &=& \left\{
   \begin{array}{ll}
   \Vec{p}_{d1}(i,j)-\Vec{p}_{d1}(i-1,j) & 1< i <W,\\
   \Vec{p}_{d1}(i,j)  & i=1,\\
   -\Vec{p}_{d1}(i-1,j)  & i=W.
   \end{array}
   \right.\\
   &+&
   \left\{
   \begin{array}{ll}
   \Vec{p}_{d2}(i,j)-\Vec{p}_{d2}(i,j-1) & 1< j <H,\\
   \Vec{p}_{d2}(i,j)  & j=1,\\
   -\Vec{p}_{d2}(i,j-1)  & j=H.
   \end{array}
   \right.
  \end{eqnarray}
  \vskip -0.1 in \noindent
\end{itemize}
Another pixel-wise estimation is the brightness $\Mat{I}_1(\Vec{x}+\Vec{u}^0)$ . It is often obtained  by warping the frame $\Mat{I}_1$ along the initial flow field $\Vec{u}^0$ using the bicubic interpolation.

\textbf{Multi-scale TV-L1.} Since the Taylor expansion is applied to linearize the brightness difference, the initial flow field $\Vec{u}^0$ should be close to the real field $\Vec{u}$ to ensure a small approximation error. To achieve this, the approximation field $\Vec{u}^0$ is derived by a multi-scale scheme in a coarse-to-fine manner. To be specific, at the coarsest level, $\Vec{u}^0$ is initialized as the zero vectors and the corresponding output of Algorithm~\ref{Alg:tvl1} is applied as the initialization of the next level\footnote{Figure 1 in the supplementary material demonstrates the framework with three-scale optimization.}.

%\begin{figure}[!h]
%\vskip -0.2 in
%\begin{center}
%\subfigure{
%\includegraphics[width=8cm]{multi-scale}
%}
%\vskip -0.1 in
%\caption{The multi-scale version of the TV-L1 method. Here the scale-size, \ie $N_{scales}$ is 3.}
%%\setlength{\belowcaptionskip}{0cm}
%\label{Fig:tvnet}
%\end{center}
%\vskip -0.2 in
%\end{figure}

\section{TVNets}
\label{Sec:tvnet}
This section presents the main contribution of this paper, \ie, the formulation of TVNet.
The central idea is to imitate the  iterative process in TV-L1 and meanwhile unfold the iterations into a layer-to-layer transformations, in the same spirit as the neural networks.

\subsection{Network design}
\label{Sec:network}
We now revisit Algorithm~\ref{Alg:tvl1} and convert its key components to a neural network.
First, the iterations in Algorithm~\ref{Alg:tvl1} can be unfolded as a fixed-size feed-forward network if we fix the number of the iterations within the while-loop to be $N_{iters}$ (see Figure~\ref{Fig:tvnet}).
%If we define each layer in the resulting network as one complete update in the while-loop, the total number of the layers is equal to $N_{scales}\times N_{warps} \times N_{iters}$.
Second, each iteration (\ie layer) is continuous and is almost everywhere smooth with respect to the input variables. Such property ensures that the gradients can be back-propagated through each layer, giving rise to an end-to-end trainable system.
%, making it possible to be applied in a globally-trainable system.

Converting Algorithm~\ref{Alg:tvl1} into a neural network involves efficiency and numerical stability considerations. To this end, we modify Algorithm~\ref{Alg:tvl1} by replacing the computations of the gradients and divergence Eq.~\eqref{Eq:central}-\eqref{Eq:backward} with specific convolutions, performing warping with bilinear interpolation, and stabilizing the division calculations with a small threshold.  We provide the details below.

\textbf{Convolutional computation.}
The most tedious part in Algorithm~\ref{Alg:tvl1} is the pixel-wise computation of Eq.~\eqref{Eq:central}-\eqref{Eq:backward}.
We propose to perform all these calculations with specific convolutional layers.
%Such property is benefit as it cancel the need to define a hand-craft layer.
We define the following kernels,
\vskip -0.3 in
\begin{eqnarray}
\label{Eq:kernels}
\Vec{w}_c = [0.5, 0, -0.5], \Vec{w}_f = \Vec{w}_b = [-1, 1].
\end{eqnarray}
\vskip -0.1 in
\noindent
Thus, for the pixels in the valid area ($1<i<W$), Eq.~\eqref{Eq:central}-\eqref{Eq:forward} can be equivalently written as
\vskip -0.25 in
\begin{eqnarray}
\label{Eq:gradients-conv}
\frac{\partial}{\partial x}\Mat{I}_1 &=&  \Mat{I}_1\ast \Vec{w}_c, \\
\frac{\partial}{\partial x}\Vec{u}_d &=&   \Vec{u}_d\ast \Vec{w}_f,
\end{eqnarray}
\vskip -0.1 in \noindent
where $\ast$ defines the convolution operation. Eq.~\eqref{Eq:kernels} only describes the kernels along the $x$ axis. We transpose them  to obtain the kernels along the $y$ axis.

The divergence in Eq.\eqref{Eq:backward} is computed by a backward difference, but the convolution is computed in a forward direction. To rewrite Eq.\eqref{Eq:backward} in convolution form, we need to first shift the pixels of $\Vec{p}_{d1}$ right (and shift $\Vec{p}_{d2}$ down) by one pixel and pad the first column of $\Vec{p}_{d1}$ (and the first row of $\Vec{p}_{d2}$) with zeros, leading to $\hat{\Vec{p}_{d1}}$ (and $\hat{\Vec{p}_{d2}}$). Then, Eq.\eqref{Eq:backward} can be transformed to
\vskip -0.25 in
\begin{eqnarray}
\label{Eq:divergence-conv}
\mathrm{div}(\Vec{p}_d) &=&   \hat{\Vec{p}_{d1}}\ast \Vec{w}_b + \hat{\Vec{p}_{d2}}\ast \Vec{w}_b^{\mathrm{T}},
\end{eqnarray}
\vskip -0.1 in \noindent
where $\Vec{w}_b^{\mathrm{T}}$ denotes the transposition of $\Vec{w}_b$.
We then refine the boundary points for the outputs of Eq.~\eqref{Eq:gradients-conv}-\eqref{Eq:divergence-conv} to meet the boundary condition in Eq.~\eqref{Eq:central}-\eqref{Eq:backward}.

\textbf{Bilinear-interpolation-based warping.}
The original TV-L1 method uses bicubic interpolation for the warping process. Here, for efficiency reasons, we adopt bilinear interpolation instead. Note that the bilinear interpolation has been applied successfully in previous works such as the spatial transformer network~\cite{jaderberg2015spatial} and the optical flow extraction method~\cite{fischer2015flownet}. We denote by $\Mat{I}_1^w = \Mat{I}_1(\Vec{x}+\Vec{u}^0)$ the warping. Then, we compute
\vskip -0.25 in
\begin{eqnarray}
\label{Eq:warping}
\nonumber \Mat{I}_1^w(i,j) &=& \sum_{n}^H\sum_m^W \Mat{I}_1(m,n)\max(0,1-|i+u_1-m|)\\
&& \max(0,1-|j+u_2-n|),
\end{eqnarray}
\vskip -0.1 in \noindent
where $u_1$ and $u_2$ are respectively the horizontal and vertical flow values of $\Vec{u}^0$ at position $(i,j)$.
We  follow the details in~\cite{jaderberg2015spatial} and derive the partial gradients for Eq.~\eqref{Eq:warping} with respect to $\Vec{u}^0$ as the bilinear interpolation is continuous and piecewise smooth.

\textbf{Numerical stabilization.}
We need to take care of the division in Algorithm~\ref{Alg:tvl1}, \ie, $\Vec{v}=-\rho(\Vec{u})\frac{\nabla\Mat{I}_1}{|\nabla\Mat{I}_1|^2}$. The operation is ill-defined when the denominator is equal to zero.
To avoid this issue, the original TV-L1 method checks whether the value of $|\nabla\Mat{I}_1|^2$ is bigger than a small constant; if not, the algorithm will set the denominator to be this small constant.
Here, we utilize a soft non-zero transformation by rewriting the update of $\Vec{v}$ as
$
\Vec{v}=-\rho(\Vec{u})\frac{\nabla\Mat{I}_1}{|\nabla\Mat{I}_1|^2+\varepsilon},
$
where a small value $\varepsilon>0$ is added to the denominator. This transformation is more efficient as we do not need to explicitly check the value of $|\nabla\Mat{I}_1|^2$ at each step.

Another division computation in Algorithm~\ref{Alg:tvl1} is
$
\label{Eq:p}
\Vec{p}_d=\frac{\Vec{p}_d+\tau/\theta \nabla\Vec{u}_d}{1+\tau/\theta|\nabla\Vec{u}_d|}.
$
At first glance, this division is safe since the denominator is guaranteed to be larger than $1$.
However,
%in next section, we will discuss how to transform TVNet to a learnable network, where back-propagating the gradients through the network is needed.
as we will see later, its gradients contain division computations where the denominators can be zero.
Thus, we apply the soft transformation by adding a small value $\varepsilon>0$ to the denominator, namely,
\vskip -0.25 in
\begin{eqnarray}
\label{Eq:new-p}
\nonumber \Vec{p}_d &=& \frac{\Vec{p}_d+\tau/\theta \nabla\Vec{u}_d}{1+\tau/\theta\sqrt{\nabla\Vec{u}_{d1}^2+\nabla\Vec{u}_{d2}^2+\varepsilon}}.\\
\end{eqnarray}
\vskip -0.1 in \noindent
The gradient of $\Vec{p}_d$ with respect to $\nabla\Vec{u}_{d1}$ is in this form
\vskip -0.25 in
\begin{eqnarray}
\label{Eq:new-gradient}
\frac{\partial}{\partial \nabla\Vec{u}_{d1}}\Vec{p}_d &=& \Vec{a}-\frac{\Vec{b}}{\sqrt{\nabla\Vec{u}_{d1}^2+\nabla\Vec{u}_{d2}^2+\varepsilon}},
\end{eqnarray}
\vskip -0.1 in \noindent
where $\Vec{a}$ and $\Vec{b}$ are well-defined variables (the details are provided in the supplementary material). In practice, both $\nabla\Vec{u}_{d1}$ and $\nabla\Vec{u}_{d2}$ are often equal to zero within the still area of the image (\eg, the background). As such, the computation of the gradients would encounter a division by zero if the positive term $\varepsilon$ was not added in Eq.~\eqref{Eq:new-gradient}. %This will cause problem for the back-propagation process which is the topic of the next section.
%Therefore, performing the soft transformation is necessary.

\textbf{Multi-scale version.} The multi-scale TVNet is formulated by directly unfolding the multi-scale version of TV-L1.
A higher scale takes as input the up-sampled output of its immediate lower scale.
There are multiple warps at each scale and each warp consists of multiple iterations.
Hence, the total number of iterations of the multi-scale TVNets are $N_{scales}\times N_{warps}\times N_{iters}$.
\subsection{Going beyond TV-L1}
\label{Sec:variable-relaxation}

In the previous section, we have transformed the TV-L1 algorithm to a feed-forward network. However, such network is parameter-free and not learnable. To formulate a more expressive network, we relax certain variables in TV-L1 to be trainable parameters. Relaxing the variables render TVNet not equivalent to TV-L1 any more. However, it allows the network to learn more complex, task-specific feature extractors by end-to-end training.

The first variable we relax is the initialization optical field $\Vec{u}^0$. In TV-L1, $\Vec{u}^0$ is set to be zero. However, from the optimization prospective, zero initialization is not necessarily the best choice; making $\Vec{u}^0$ trainable will enable us to automatically determine a better initialization for the optimization.
We also propose to relax the convolutional filters in Eq.~\eqref{Eq:gradients-conv}-\eqref{Eq:divergence-conv}. The original convolutions are used to derive the (numerical) gradients and divergences.
Allowing the convolutional filters to be trainable parameters will enable them to discover more complex patterns in a data-driven way. We will demonstrate the benefit of the trainable version compared to the original architecture in our experiments.

\subsection{Multi-task Loss}

As discussed before, our TVNet can be concatenated to any task-specific networks (\eg, the BN-Inception net for action classification~\cite{Wang_ECCV16}) to perform end-to-end action recogntion without the need of explicitly extracting the optical flow data, as illustrated in Figure~\ref{Fig:tvnet} (c).
Because of the end-to-end structure, the parameters of TVNet can be fine-tuned by back-propagating gradients of the task-specific loss.
Additionally, since the original TV-L1 method is developed to minimize the energy function in Eq.~\eqref{Eq:energy-function}, we can also use this function as an additional loss function to force it to produce meaningful optical-flow-like features. To this end, we formulate a multi-task loss as
\vskip -0.25 in
\begin{eqnarray}
\label{Eq:loss}
L = L_{c} + \lambda L_{f}.
\end{eqnarray}
\vskip -0.1 in \noindent
Here $L_{c}$ is the action classification loss (\eg the cross entropy), $L_{f}$ is defined in Eq.~\eqref{Eq:energy-function} where the exact computation other than the Tailor approximation is applied to compute $\Vec{\rho}(\Vec{u}(\Vec{x}))$, and $\lambda$ is a hyper-parameter to trade-off these two losses. We  set $\lambda=0.1$ in all our experiments and find that it works well across all of them.
Note that it is tractable to compute the gradients of $L_{f}$ as it has been translated to convolutions and the bilinear interpolation (see \textsection~\ref{Sec:network}).
%that such value derives desired performance.
%Note that the loss $L_{f}$ is unsupervised in the sense that it does not need any ground true of flow fields, thus making our network clearly different from those architectures using CNNs to learn optical flow in a supervised way [].

\section{Experiments}

This section performs experimental evaluations to verify the effectiveness of the proposed TVNet. We first carry out a complete comparison between TVNets of various structures with the TV-L1 method regarding the optimization efficiency.
Then, we compare the performance of TVNets with state-of-the-art methods on the task of action recognition\footnote{We also provide additional experimental evaluations on action similarity labeling in the supplementary material.}. %For both tasks, the performance relies on the action representation ability. A better action representor enables a better accuracy.

The three hyper-parameters, $N_{scales}$, $N_{warps}$ and $N_{iters}$ determine the structure of the TVNet.
For convenience, we denote the TVNet with particular values of the hyper-parameters as TVNet-$N_{scales}$-$N_{warps}$-$N_{iters}$. We denote the architecture as TVNet-$N_{iters}$ for short when both $N_{scales}$ and $N_{warps}$ are fixed to be 1.
For the TV-L1 method, the hyper-parameters are fixed as $N_{scales}=N_{warps}=5$ and $N_{iters}=50$ in all experiments unless otherwise specified.
Our methods are implemented by the Tensorflow platform~\cite{abadi2016tensorflow}. Unless otherwise specified, all experiments were performed on 8 Tesla P40 GPUs.

\subsection{Comparison with TV-L1}

Initialized as a particular TV-L1 method, the parameters of TVNet can be further finetuned as discussed in Section~\ref{Sec:variable-relaxation}. Therefore, it is interesting to evaluate how much the training process can improve the final performance. For this purpose, we compare the estimation errors between TVNet and TV-L1 on the optical flow dataset, \ie, the MiddleBurry dataset~\cite{baker2011database}.

\begin{table}[t!]
\vskip -0.2 in
\centering
\caption{The average EPEs on MiddleBurry. ``Training $\Vec{u}^0$'' means only $\Vec{u}^0$ is trained; ``All Training'' means both $\Vec{u}^0$ and the convolution filters are trained. After training, TVNet-50 outperforms TV-L1 significantly although TV-L1 has a much larger number of optimization iterations (\ie, 1250).}
\label{Tab:vs-tvl1}
%\begin{tabular}{c|c|c}
%\toprule
%Methods         & Before training & After training \\
%\hline
%TVNet-10        & 3.47     &  1.24   \\
%TVNet-30        & 3.01     &  0.40  \\
%TVNet-3-1-10    & 2.00     &  0.52 \\
%TVNet-1-3-10    & 2.81     &  0.46   \\
%TVNet-50        & 2.93     &  \textbf{0.35}    \\
%\hline
%TV-L1           & \multicolumn{2}{c}{0.66}          \\
%\bottomrule
%\end{tabular}
%\vskip -0.2 in
%\end{table}
\tabcolsep 4pt %space between two columns.
\renewcommand{\arraystretch}{0.8}
\begin{tabular}{c|c|c|c}
\toprule
Methods         & No training & Training $\Vec{u}^0$ & All Training \\
\hline
TVNet-10        & 3.47  & 2.92  &  1.24   \\
TVNet-30        & 3.01  & 2.04  &  0.40  \\
TVNet-3-1-10    & 2.00  & 0.82  &  0.52 \\
TVNet-1-3-10    & 2.81  & 2.17  &  0.46   \\
TVNet-50        & 2.93  & 1.58  &  \textbf{0.35}    \\
\toprule
\toprule
TV-L1-10   & 3.48   &  TV-L1-3-1-10    & 1.79        \\
TV-L1-30   & 3.02   &  TV-L1-1-3-10    & 2.74     \\
TV-L1-50   & 2.86   &  TV-L1-5-5-50    & 0.66   \\
\bottomrule
\end{tabular}
\vskip -0.05in
\end{table}

\begin{figure}[t!]
\vskip -0.1 in
\begin{center}
\subfigure{
\includegraphics[width=\columnwidth]{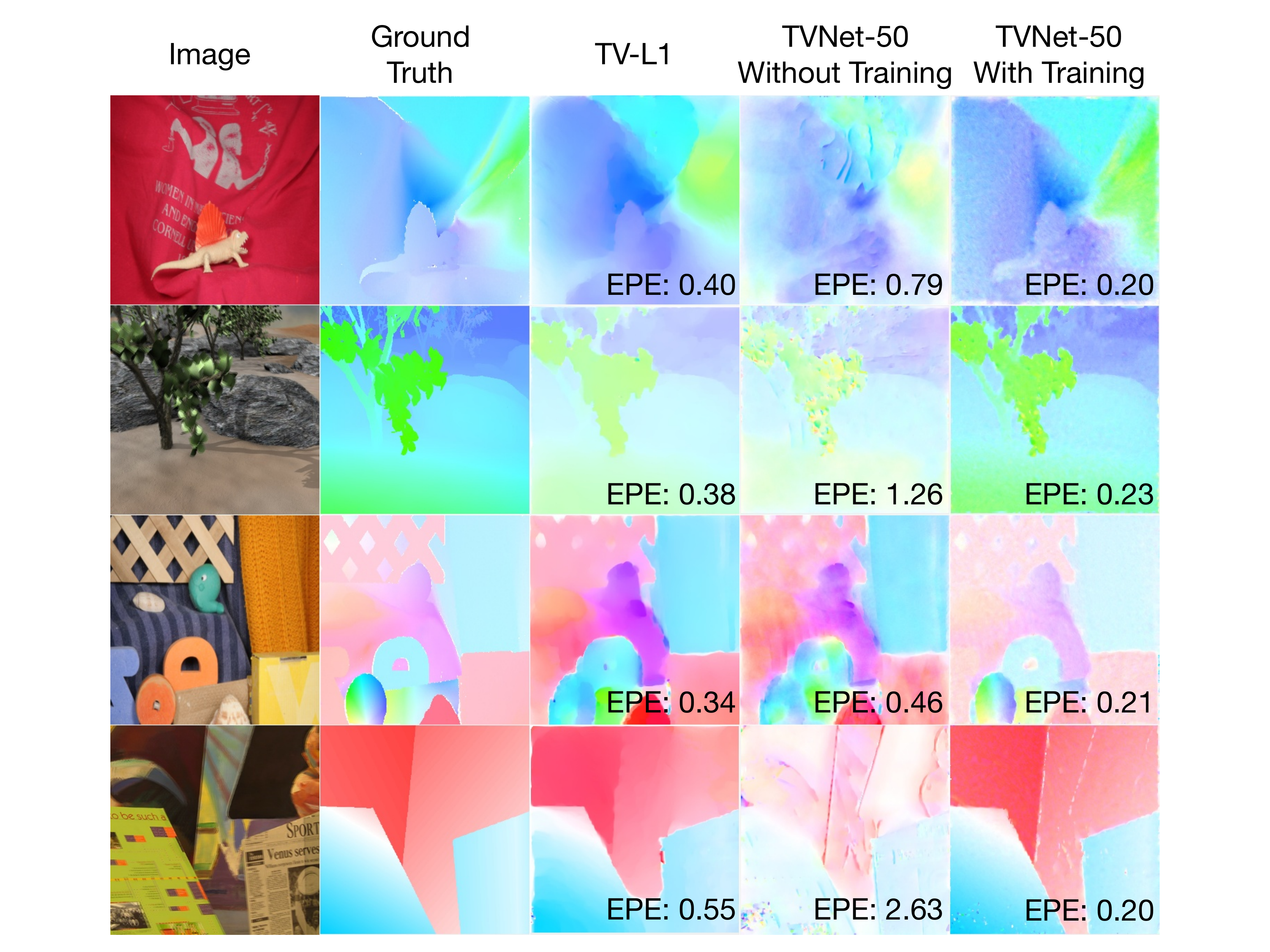}
}
\vskip -0.1 in
\caption{Examples of flow fields from TV-L1 and TVNet-50 estimated on MiddleBurry. With training, TVNet-50 is able to
extract finer details than TV-L1 does.}
\label{Fig:flow}
\end{center}
%\vskip -0.1 in
\end{figure}

\textbf{Dataset.}
The MiddleBurry dataset~\cite{baker2011database} is a widely-used benchmark for evaluating different optical flow extraction methods.
Here we only perform evaluation on the training set as we are merely concerned about the training efficiency of TVNets.
%This dataset consists of one training and one evaluation sets, each of which contains 12 image sequences in total.
For the training set, only 8 image pairs are provided with the ground-true optical flow.

\textbf{Implementation details.}
The estimation errors are measured via the average End-Point Error (EPE) defined by
\begin{eqnarray}
\label{Eq:EPE}
EPE\doteq \frac{1}{N}\sum_{i=1}^N\sqrt{(u_{1,i}-u_{1,i}^{gt})^2+(u_{2,i}-u_{2,i}^{gt})^2},
\end{eqnarray}
where $(u_{1,i},u_{2,i})$ and $(u_{1,i}^{gt},u_{2,i}^{gt})$ are the predicted and ground-true flow fields, respectively.
For the training of TVNets, we adopt the EPE (Eq.~\eqref{Eq:EPE}) as the loss function, and apply the batch gradient decent method with the learning rate and max-iteration being 0.05 and 3000, respectively.
Several structures, \ie, TVNet-10, TVNet-30, TVNet-50, TVNet-3-10, TVNet-1-3-10, and their counterparts of TV-L1 are implemented to compare the difference between different network designs.

\begin{table}[t!]
\vskip -0.2 in
\centering
\caption{The execution speed of different flow extraction methods. Only one gpu is used for the evaluations. As no ground-truth is given on UCF101, we apply the term $\rho(\Vec{u})$ (Eq.(1)) instead of End-Point-Error to measure the optical flow error. TVNet-50 achieves the fastest speed among all the methods. Theoretically, since TVNet-50 has a much smaller number of iterations than TV-L1 (\ie 50 v.s. 1250), the speed of TVNet-50 should be more than 100 times faster than TV-L1. However, due to the different implementations of TV-L1 and TVNet, the real computational reduction of TVNet is not so big.
As the TVNet-50 is implemented by Tensorflow, we can easily perform parallel flow extraction with TVNet-50 by enlarging the batch size of input images (\eg, setting batch = 10); as such, the FPS will be further improved to 60.}
\label{Tab:time}
%\begin{tabular}{c|c}
%\toprule
%Methods         & frames per second \\
%\hline
%DIS-Fast        & 9.23         \\
%Deepflow        & 0.69   \\
%Flownet2.0      & 4.53         \\
%TV-L1           & 6.67         \\
%\hline
%TVNet-50 (batch=1)       & \textbf{12}  \\
%TVNet-50 (batch=10)      & \textbf{60}  \\
%\bottomrule
%\end{tabular}
%\end{table}
\tabcolsep 4pt %space between two columns.
\renewcommand{\arraystretch}{0.8}
\begin{tabular}{c|c|c|c|c}
\toprule
Methods         & FPS  & Flow Errors & Trainable & \#Parameters\\
\hline
DIS-Fast        & 9.23  & 1.29 & No & No     \\
Deepflow        & 0.69  & 1.33 & No & No     \\
Flownet2.0      & 4.53  & 1.32 & Yes & $10^5$     \\
TV-L1           & 6.67  & 0.86 & No & No     \\
\hline
TVNet-50       & \textbf{12} & 0.93 & Yes &$10^2$ \\
%TVNet-50 (batch=10)      & \textbf{60}  & &\\
\bottomrule
\end{tabular}
\vskip -0.05in
\end{table}

\begin{table}[t!]
%\vskip -0.1 in
\centering
\caption{Classification accuracy of various motion descriptors on HMDB51 (split 1) and UCF101 (split 1).The top part shows the results of current best action representation methods; the middle part reports the accuracies of the four baselines; the bottom part presents the performance of our models. TVNet-50 achieves the best results on both datasets.}
\label{Tab:action}
\tabcolsep 4pt %space between two columns.
\renewcommand{\arraystretch}{0.8}
\begin{tabular}{c|c|c}
\toprule
Methods                       & HMDB51         & UCF101                    \\
\hline
C3D~\cite{tran2015learning}                         & -              & 82.3\%                      \\
ActionFlowNet~\cite{ng2016actionflownet}              & 56.4\%         & 83.9\% \\
\hline
TV-L1                         & 56.0\%         & 85.1\%                     \\
DIS-Fast                      & 40.4\%         & 71.2\%                \\
Deepflow                      & 50.4\%         & 82.1\%                    \\
Flownet2.0                    & 52.3\%         & 80.1\%                    \\
\hline
TVNet-50 (no training)        & 55.6\%         & 83.5\%                \\
TVNet-50 (no flow loss)       & 56.9\%         & 84.5\%                    \\
TVNet-50                      & \textbf{57.5}\%& \textbf{85.5}\%                       \\
\bottomrule
\end{tabular}
\vskip -0.1 in
\end{table}

\textbf{Results.}
We have performed one-to-one comparisons between TVNets and TV-L1 on MiddleBurry in Table~\ref{Tab:vs-tvl1}. Given the same architecture, TVNet without training achieves close performance to TV-L1. This is not surprising since TVNet and TV-L1 are almost the same except the way of interpolation (bilinear vs.\ bicubic). To further evaluate the effect of training $\Vec{u}^0$, we conduct additional experiments and report the results in Table~\ref{Tab:vs-tvl1}. Clearly, making $\Vec{u}^0$ trainable in TVNets can indeed reduce the End-Point Error.
With training both $\Vec{u}^0$ and the convolution filters, all TVNets except TVNet-10 achieve lower errors than TV-L1-5-5-50, even though the number of iterations in TVNets (not more than 50) are much smaller than that of TV-L1-5-5-50 (up to 1250).
Figure~\ref{Fig:flow} displays the visualization of the optical flow between TV-L1-5-5-50 and TVNet-50.
%The training process enables the huge improvement for the networks even with small size.
Another interesting observation is from the comparison between  TVNet-30, TVNet-50, TVNet-3-10 and TVNet-1-3-10. It is observed TVNet-30 and TVNet-50 finally outperform TVNet-3-10 and TVNet-1-3-10 after training,  implying that the flat structure (\ie $N_{scales}=N_{warps}=1$) is somehow easier to train.
For the remaining experiments below, we will only compare the performance between TVNet-50 and TV-L1-5-5-50, and denote TV-L1-5-5-50 as TV-L1 for similarity.

%\begin{table*}[]
%\centering
%\caption{My caption}
%\label{my-label}
%\begin{tabular}{c|c|c|c|c|c|c|c|c}
%\toprule
%\multirow{2}{*}{Methods} & \multirow{2}{*}{DIC-Fast} & \multirow{2}{*}{DeepFlow} & \multirow{2}{*}{FlowNet2} & \multirow{2}{*}{TV-L1} & \multicolumn{2}{c|}{unsuprvised} & \multicolumn{2}{c}{surpervise} \\ \cline{6-9}
%                         &                           &                           &                           &                        & TVNet-50    & TVNet-Inception    & TVNet-50    & TVNet-Inception   \\ \hline
%   EPE  &        0.25   &   0.92  &  0.35   &    0.90     &             &                    &      0.39       &                   \\ \bottomrule
%\end{tabular}
%\end{table*}

\subsection{Action recognition}

\begin{figure*}[th!]
%\vskip -0.2 in
\begin{center}
\subfigure{
\includegraphics[width=14cm,height=6cm]{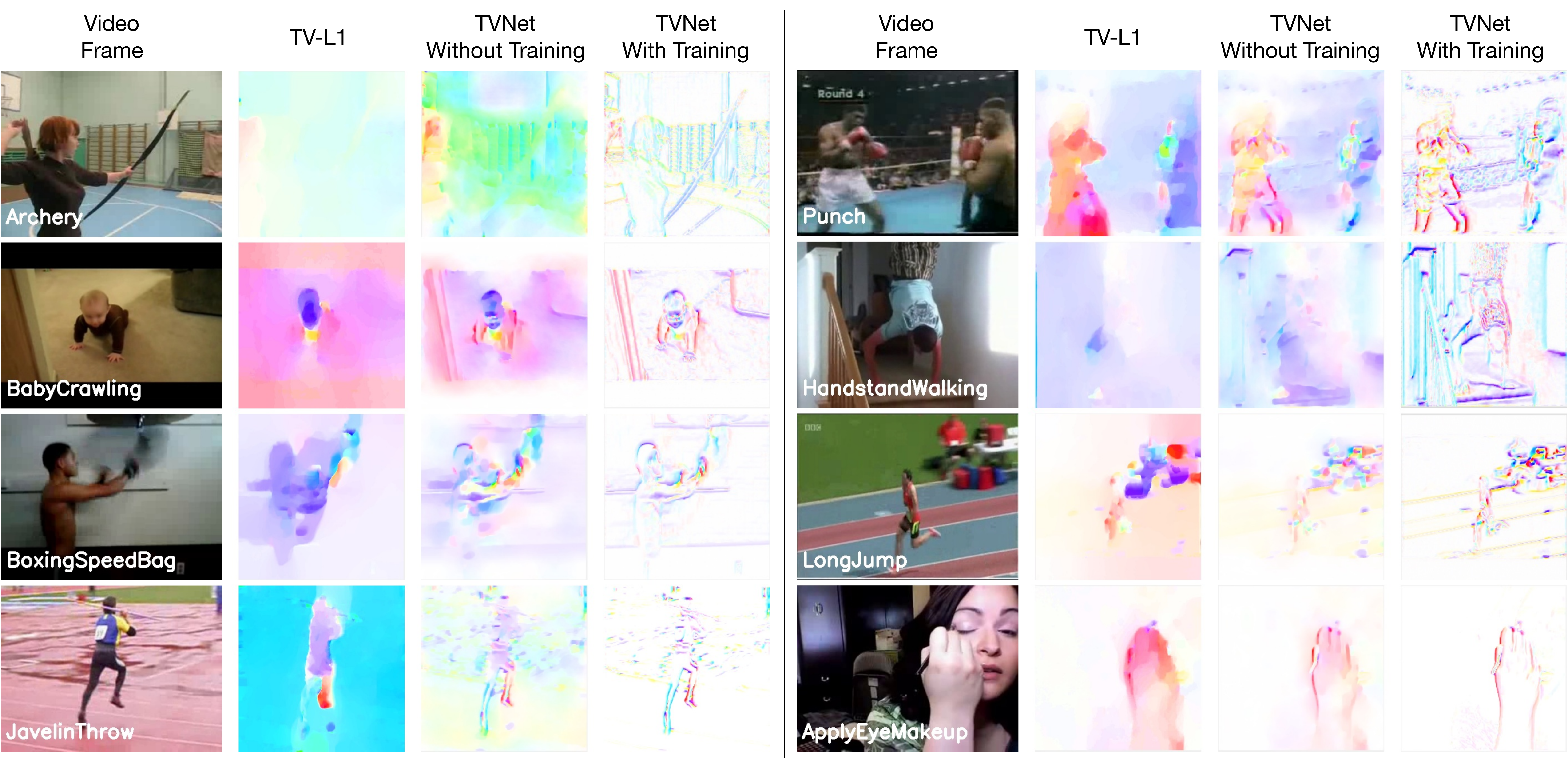}
}
\vskip -0.05 in
\caption{Illustrations of the motion patterns obtained by TV-L1 and TVNet-50 on the UCF101 dataset. From the first to the last column, we display the image-pair (first image only), the motion features by TV-L1, TVNet-50 without training and with training, respectively. Interestingly, with training, TVNet-50 generates more abstractive motion features than TV-L1 and its non-trained version. These features not only automatically remove the movement of the background (see the ``punch'' example), but also capture the outline of the moving objects.}
\label{Fig:representation}
\end{center}
\vskip -0.2 in
\end{figure*}

\textbf{Dataset.}
Our experiments are conducted on two popular action recognition datasets, namely the {UCF101}~\cite{soomro2012ucf101} and
the {HMDB51}~\cite{kuehne2011hmdb} datasets.  The {UCF101} dataset contains 13320 videos of 101 action classes. The {HMDB51} dataset consists of 6766 videos from 51 action categories.
%The videos in the {HMDB51} dataset are collected from various sources, such as movies and web videos.
%For both datasets, we follow the provided evaluation protocols and report the average accuracy over the three training/testing splits.

\textbf{Implementation details.}
As discussed before, our TVNets can be concatenated by a classification network to formulate an end-to-end model to perform action recognition. We apply the BN-Inception network~\cite{Wang_ECCV16} as the classification model in our experiments due to its effectiveness. The BN-Inception network is pretrained by the cross-modality skill introduced in~\cite{wang2015towards} for initialization.

We sample a stack of 6 consecutive images from each video and extract 5 flow frames for every consecutive pair. The resulting stack of optical flows are fed to the BN-Inception network for prediction. To train the end-to-end model, we set the mini-batch size of the sampled stacks to 128 and the momentum to 0.9.
The learning rate was initialized to 0.005. The maximum number of learning iterations for the {UCF101} and the {HMDB51}
datasets was chosen as 18000 and 7000, respectively. We decreased the learning rates by a factor of 10 after
the 10000th and 16000th iterations for the {UCF101} experiment, and after 4000th and 6000th iterations for the {HMDB51} case. We only implement TVNet-50 in this experiment. To prevent overfitting, we also carry out the corner cropping and scale jittering~\cite{Wang_ECCV16}; the learning rate for TVNets is further divided by 255.

For the testing, stacks of flow fields are extracted from the center and four corners of a video. We sample 25 stacks from each location (\ie, center and corners), followed by flipping them horizontally to enlarge the testing samples. All the sampled snippets (250 in total) are fed to BN-Inception~\cite{Wang_ECCV16} and their outputs are averaged for prediction.

\textbf{Baselines.}
Beside the TV-L1 method, we carry out other three widely-used flow extraction baselines including DIS-Fast~\cite{kroeger2016fast}, DeepFlow~\cite{weinzaepfel2013deepflow} and FlowNet2.0~\cite{ilg2016flownet}. For FlowNet2.0, we use the pretrained model by the KITTI dataset.  For all baselines, we compute the optical flow beforehand and store the flow fields as JPEG images by linear compression. All methods share the same training setting and classification network for fair comparison.

\begin{table}[t!]
%\vskip -0.2 in
\centering
\caption{Mean classification accuracy of the state-of-the-arts on HMDB51 and UCF101.}
\label{Tab:without-IDT}
\renewcommand{\arraystretch}{0.8}
\tabcolsep 8pt %space between two columns. ÓÃÓÚµ÷ÕûÁÐ¼ä¾à
\begin{tabular}{ccc}
\toprule
  Method                                       & HMDB51      & UCF101      \\ \hline
  ST-ResNet~\cite{Feichtenhofer_NIPS16}        & 66.4\%      &  93.4\%     \\
  ST-ResNet + IDT~\cite{Feichtenhofer_NIPS16}                           & 70.3\%       & 94.6\%     \\
  TSN~\cite{Wang_ECCV16}                       & 68.5\%       &   94.0\%   \\
  KVMF~\cite{Wangjiang}                        &  63.3\%     & 93.1\%      \\
  TDD~\cite{wang2015action}              & 65.9\%       & 91.5\%     \\
  %Dynamic Image Networks + IDT~\cite{bilen2016dynamic}                  & 89.1\%      & 65.2\%      \\
%  Two-Stream Fusion + IDT~\cite{Feichtenhofer_CVPR16}                   & 93.5\%      & 69.2\%      \\
  C3D (3 nets)~\cite{tran2015learning}            & -     &  90.4\%           \\
  Two-Stream Fusion~\cite{Feichtenhofer_CVPR16}& 65.4\%      & 92.5\%      \\
  Two-Stream (VGG16)~\cite{ballas2015delving}  & 58.5\%      & 91.4\%     \\
  Two-Stream+LSTM~\cite{Ng_CVPR15}             &  -          & 88.6\%      \\
  Two-Stream Model~\cite{Simonyan_NIPS14}      & 59.4\%      & 88.0\%      \\

  \hline
%  TVNet-50 + RGB                                & 90.1\%      &   65.0\%    \\
  Ours                           & 71.0\%      &  94.5\%     \\
  Ours + IDT                     & \textbf{72.6}\%      &  \textbf{95.4}\%     \\
  \bottomrule
\end{tabular}
\vskip -0.2 in
\end{table}

\textbf{Computational efficiency comparisons.}
We have added thorough computational comparison between TVNet, TV-L1, DIS-Fast, Deepflow, and Flownet2.0 in Table~\ref{Tab:time}.  To do so, we randomly choose one testing video from the UCF101 dataset, and compute the optical flow for every two consecutive frames. The average running time (excluding I/O times) for TVNet-50, TV-L1, DIS-Fast, DeepFlow and FlowNet2.0 are summarized in Table~\ref{Tab:time}.
The results verify the advantages of TVNets regarding high number of Frames-per-Second (FPS), low optical flow error, end-to-end trainable property, and small number of model parameters.
Flownet2.0 performs more accurately than TV-L1 on the optical flow datasets (e.g. MiddleBurry) as reported by~\cite{ilg2016flownet}. However,
for the action datasets, TV-L1 and our TVNet obtain lower flow error than Flownet2.0 according to Table~\ref{Tab:time}.

\textbf{Classification accuracy comparisons.}
Table~\ref{Tab:action} presents action recognition accuracies of TVNet-50 compared with the four baselines and current best action representation methods.
Clearly, TVNet-50 outperforms all compared methods on both datasets. Compared to TV-L1, the improvement of TVNet-50 on UCF101 is not big; however, our TVNet-50 is computationally advantageous over TV-L1 because it only employs one scale and one warp, while TV-L1 adopts five scales and five warps.  Even when we freeze its parameters, TVNet-50 still achieves better results than DIS-Fast, DeepFlow and FlowNet2.0; as our TVNet is initialized as a special TV-L1, the initial structure is sufficient to perform promisingly.
%By enabling the training of TVNet-50, the accuracies are further increased by 2\% on UCF101 and 1.9\% on HMDB51.
The TVNet is trained with the multi-task loss given by Eq.~\eqref{Eq:loss}. To verify the effect of the flow loss term, \ie $L_f$,
we train a new model only with the classification loss. Table~\ref{Tab:action} shows that such setting  decreases the accuracy.

Flownet2.0 can also be jointly finetuned for action classification. This is done in ActionFlowNet~\cite{ng2016actionflownet} but the results, as incorporated in Table~\ref{Tab:action}, are worse than ours. This is probably because TVNet has much fewer parameters than Flownet2.0, making the training more efficient and less prone to overfitting.
For the UCF101 dataset, the TVNet outperforms C3D~\cite{tran2015learning} by more than $2\%$. The C3D method applied 3-dimensional convolutions to learn spatiotemporal features. In contrast to this implicit modeling, in our model, the motion pattern is extracted by TVNet explicitly.
We also visualize the outputs by TV-L1 and TVNets in Figure~\ref{Fig:representation}.
%Interestingly, after task-specific training, TVNet-50 generates more abstractive motion features than those of TV-L1 and its non-trainable version. These features not only automatically remove the movement of the background (\ie the dancing example), but also capture the outline of the moving objects.
%We also visualize their activations by DeepDraw\footnote{https://github.com/auduno/deepdraw} in Figure~\ref{Fig:representation}. It is observed that the features generated by TVNet-50 mostly capture the movement patterns.

\textbf{Comparison with other state-of-the-arts.}
To compare with state-of-the-art methods, we apply several practical tricks to our TVNet-50, as suggested by previous works~\cite{Simonyan_NIPS14,Wang_ECCV16}. First, we perform the two-stream combination trick~\cite{Simonyan_NIPS14} by additionally training a spatial network on RGB images. We use the BN-Inception network as the spatial network and apply the same experimental setting as those in~\cite{Wang_ECCV16} for the training. At testing, we combine the predictions of the spatial and temporal networks with a fixed weight (\ie, 1:2). Second, to take the long-term temporal awareness into account, we perform the temporal pooling of 3 sampled segments for each video during training as suggested by~\cite{Wang_ECCV16}.

Table~\ref{Tab:without-IDT} summarizes the classification accuracy of TVNets compared with the state-of-the-art approaches over all three splits of the {UCF101} and the {HMDB51} dataset.
The improvements achieved by TVNets are quite substantial compared to the original two-stream method~\cite{Simonyan_NIPS14} (6.5\% on {UCF101} and 11.6\% on  {HMDB51}). Such significant gains are achieved as a result of employing better models (\ie, BN-Inception net)
and also considering end-to-end motion mining.

The TSN method~\cite{Wang_ECCV16} is actually a two-stream model with TV-L1 inputs.
TSN shares the same classification network and experimental setups as our TVNets.
As shown in Table~\ref{Tab:action}, our TVNets outperform TSN on both action datasets (e.g. 71.6\% vs.\ 68.5\% on HMDB51), verifying the effectiveness of TVNets for the two-stream models.
%The Spatial-Temporal-Residue-Network (ST-ResNet) achieves the most close result to our method. This approache is benefited from a very deep network (\ie, 50-layer ResNet) along cross-stream residual connections~\cite{Feichtenhofer_NIPS16}. However, its accuracies are more than 1\% lower than our results on both datasets.

Combining CNN models with trajectory-based hand-crafted IDT features~\cite{wang2013action} can improve the final performances ~\cite{wang2015action,tran2015learning,bilen2016dynamic,Feichtenhofer_NIPS16}. Hence, we averaged the L2-normalized SVM scores of FV-encoded IDT features (\ie, HOG, HOF and MBH) with the L2-normalized video predictions (before the loss layer) of our methods. Table~\ref{Tab:without-IDT} summarizes the results and indicates that there is still room for improvement. Our 95.4\% on the {UCF101} and 72.6\% on the {HMDB51} remarkably outperform all the compared methods.

A recent state-of-the-art result is obtained by I3D~\cite{carreira2017quo}, achieving 97.9\% on {UCF101} and 80.2\% on {HMDB51}. However, the I3D method improves the performance by using a large amount of additional training data. It is unfair to compare their results with ours.

\section{Conclusion}
In this paper, we propose a novel end-to-end motion representation learning framework, named as TVNet. Particularly, we formulate the TV-L1 approach as a neural network, which takes as input stacked frames and outputs optical-flow-like motion features. Experimental results on two video understanding tasks demonstrate its superior performances over the existing motion representation learning approaches. In the future, we will explore more large-scale video understanding tasks to examine the benefits of the end-to-end motion learning method. %We hope this paper will open up avenues for exploitation of end-to-end motion learning with deep networks for various video understanding tasks.
%\begin{eqnarray}
%\label{Eq:p}
%\nonumber \Vec{p}_d &=& \frac{\Vec{p}_d+\tau/\theta \nabla\Vec{u}_d}{1+\tau/\theta|\nabla\Vec{u}_d|},\\
%&=& \frac{\Vec{p}_d+\tau/\theta \nabla\Vec{u}_d}{1+\tau/\theta\sqrt{\nabla\Vec{u}_{d1}^2+\nabla\Vec{u}_{d2}^2}}.
%\end{eqnarray}
%In next section, we will discuss how to transform TVNet to a learnable network. As such, we need to back-propagate the gradients through each update.
%If we want to compute the gradient of $\Vec{p}_d$ with respect to $\nabla\Vec{u}_{d1}$, we achieve
%\begin{eqnarray}
%\label{Eq:gradient}
%\frac{\partial}{\partial \nabla\Vec{u}_{d1}}\Vec{p}_d &=& \Mat{\Gamma}-\frac{\alpha\nabla\Vec{u}_{d1}}{\sqrt{\nabla\Vec{u}_{d1}^2+\nabla\Vec{u}_{d2}^2}},
%\end{eqnarray}
%where $\Mat{\Gamma}$ and $\alpha$ are well-defined variables (see the supplementary material).

%\begin{eqnarray}
%\label{Eq:gradient}
%\nonumber \frac{\partial}{\partial \nabla\Vec{u}_{d1}}\Vec{p}_d &=& \frac{\tau/\theta }{(1+\tau/\theta|\nabla\Vec{u}_d|)^2}((1+\tau/\theta|\nabla\Vec{u}_d|)\frac{\partial}{\partial \nabla\Vec{u}_{d1}}\nabla\Vec{u}_d\\
%&& -\frac{\nabla\Vec{u}_{d1}}{\sqrt{\nabla\Vec{u}_{d1}^2+\nabla\Vec{u}_{d2}^2}}.
%\end{eqnarray}

{\small
\bibliographystyle{ieee}
\bibliography{reference}
}

\end{document}